%% file: main.tex
\documentclass[letterpaper, 10 pt, conference]{ieeeconf}
\IEEEoverridecommandlockouts
\usepackage{caption}
\usepackage[utf8]{inputenc} 

\usepackage[T1]{fontenc}    
\usepackage{hyperref}       
\usepackage{url}            
\usepackage{booktabs}       
\usepackage{amsfonts}       
\usepackage{nicefrac}       
\usepackage{microtype}      
\usepackage{xcolor}         
\usepackage{multirow}
\usepackage{graphicx}
\usepackage{amsmath} 
\usepackage{makecell}

\def\etal{\emph{et al}}

\newcommand{\net}{AVT-SSDepth}

\begin{document}

\title{\LARGE \bf Adjacent-view Transformers for Supervised Surround-view \\ Depth Estimation}

\author{
\vspace{-5mm}
 \\
  Xianda Guo$^{1*}$~~~~~~~
  Wenjie Yuan$^{2*}$~~~~~~~
  Yunpeng Zhang$^{3}$~~~~~~~
  Tian Yang$^{4}$~~~~~~~
  Chenming Zhang$^{5}$~~~~~~~\\
  Zheng Zhu$^{6,\dagger}$~~~~~~~ 
  Qin Zou$^{1,\ddagger}$~~~~~~~
  Long Chen$^{7,5,2,\ddagger}$~~~~~~~\\
  \textsuperscript{1} School of Computer Science, Wuhan University~~~~\textsuperscript{2}Waytous~~~~\textsuperscript{3} Tsinghua University\\
  \textsuperscript{4} PhiGent Robotics~~~~
  \textsuperscript{5} Institute of Artificial Intelligence and Robotics, Xi'an Jiaotong University\\
  \textsuperscript{6} GigaAI
  \textsuperscript{7} Institute of Automation, Chinese Academy of Sciences~~~~
  \\
\texttt{xianda\_guo@163.com; wenjie\_yuan@outlook.com; long.chen@ia.ac.cn}
}



\maketitle

\thispagestyle{empty} 
\pagestyle{empty}  

\renewcommand{\thefootnote}{\fnsymbol{footnote}}
\footnotetext[1]{These authors contributed to the work equally.}
\footnotetext[2]{Project Leader.}, \footnotetext[3]{Corresponding authors.}

\input{IROS/0abstract}
\input{IROS/1introduction}

\input{IROS/2relatedwork}

\input{IROS/3method}
\input{IROS/4experients}
\input{IROS/5conclusion}


\bibliographystyle{IEEEtran}
\bibliography{egbib}

\end{document}

%% file: IROS/0abstract.tex
\begin{abstract}
Depth estimation has been widely studied and serves as the fundamental step of 3D perception for robotics and autonomous driving. Though significant progress has been made in monocular depth estimation in the past decades, these attempts are mainly conducted on the KITTI benchmark with only front-view cameras, which ignores the correlations across surround-view cameras. In this paper, we propose an \textbf{A}djacent-\textbf{V}iew \textbf{T}ransformer for \textbf{S}upervised \textbf{S}urround-view \textbf{Depth} estimation (\textbf{\net{}}), to jointly predict the depth maps across multiple surrounding cameras. Specifically, we employ a global-to-local feature extraction module that combines CNN with transformer layers for enriched representations. Further, the adjacent-view attention mechanism is proposed to enable the intra-view and inter-view feature propagation. The former is achieved by the self-attention module within each view, while the latter is realized by the adjacent attention module, which computes the attention across multi-cameras to exchange the multi-scale representations across surround-view feature maps. In addition, \textbf{\net{}} has strong cross-dataset generaliza-
tion. Extensive experiments show that our method achieves superior performance over existing state-of-the-art methods on both DDAD and nuScenes datasets.  Code is available at \url{https://github.com/XiandaGuo/SSDepth}.
\end{abstract}

%% file: IROS/1introduction.tex
\section{Introduction}
\label{sec:intro}
With the rapid development of robotics and autonomous driving technology in recent years, depth estimation has received increasingly more attention. As a bridge connecting the 2D images and the 3D environments, depth estimation aims to recover the dense 3D representations from 2D images. 
Monocular depth estimation~\cite{eigen2014depth,monodepth2, monovit, duan2023diffusiondepth} aims to get the depth map of a single image, which either uses LiDAR points as the target depth or learns from monocular videos in a self-supervised manner. Although significant progress has been made for monocular depth estimation in the past decade, existing methods are mainly performed on only a few driving scenes, mostly KITTI datasets~\cite{kitti}. 
Stereo matching~\cite{chang2018pyramid, guo2023openstereo, guo2024lightstereo} attempts to predict the disparity between the left and right camera images, where a large overlap between the stereo images is assumed. 
Though these methods achieve impressive performance for depth estimation, their receptive fields are limited to a small fraction of the entire 3D environment (usually less than $180^\circ$). 

Although recently released datasets contain consistent multi-camera data that capture the full $360^\circ$ point cloud~\cite{3Dpacknet,caesar2020nuscenes}, existing methods still handle the problems as monocular~\cite{PixelFormer,eigen2015predicting, BTS,yin2019enforcing,fu2018deep} and ignore the multi-view interaction. Multi-View Stereo~\cite{mvsnet,xue2019mvscrf,wei2021nerfingmvs,xue2019mvscrf,huang2018deepmvs} is an effective method to predict a depth map by a series of calibrated images surrounding an object, but it needs large overlapping areas between cameras.
FSM~\cite{guizilini2021full} firstly extends self-supervised learning from monocular and stereo to the general multi-camera setting and introduces spatial-temporal contexts and pose consistency constraints, which boost performance in multi-camera settings. Following FSM~\cite{guizilini2021full}, SurroundDepth~\cite{wei2022surround} proposes to jointly process all surrounding views by the module of cross-view attention to get high-quality depth maps. In addition, SurroundDepth~\cite{wei2022surround} can get the depths with real-world scales by the scale-aware SfM pretraining. However, the performance of these methods is insufficient for practical applications.

Multi-view 3D object detection in Bird’s-Eye-View (BEV) has received great attention recently because it has a much lower computational cost for autonomous driving than LiDAR-based solutions. Estimating accurate depth is essential for multi-view 3D object detection. However, it remains challenging to predict depth in surrounding images without overlapping areas. BEVDepth~\cite{BEVDepth} provides a quantitative and qualitative analysis of depth estimation within 3D detectors and boosts the performance of 3D detectors by using ground-truth depth. 
Although these methods have explored the effect of depth estimation on BEV 3D object detection, there is no baseline for supervised surround-view depth estimation.
Therefore, it is time to build a baseline for supervised surround-view depth estimation.

\input{figures_tex/arch}

In this paper, we propose \textbf{A}djacent-\textbf{V}iew \textbf{T}ransformers \textbf{S}upervised \textbf{S}urround-view \textbf{Depth} estimation, called \textbf{\net{}}. Specifically, we first employ a shared encoder that can extract global-to-local features by combining CNN with transformer layers. To learn adjacent-view interactions in surrounding views, we adopt the alternating attention mechanism: self-attention computes attention in the same view, while adjacent attention computes attention between adjacent views. Finally, different from monocular depth estimation, a Depth Head is used to predict all surround-view depth maps at the same time. We conduct extensive experiments on the challenging datasets DDAD \cite{3Dpacknet} and nuScenes \cite{caesar2020nuscenes} for multi-camera depth estimation.

In summary, the main contributions are as follows:

\begin{itemize}
    \item For the first time, we propose adjacent-view transformers for supervised surround-view depth estimation to jointly predict depth maps in a wide-baseline $360^\circ$ multi-camera setting.   
    \item We use the joint convolutional attention and transformer block for surround-view depth estimation to model local and global contexts. Furthermore, we propose \textbf{adjacent-view attention} mechanism, which can be used to enable the intra-view and inter-view feature propagation in this novel setting.
    \item Our \net{} yields substantial improvements to depth estimation when compared to state-of-the-art methods on DDAD~\cite{3Dpacknet} and nuScenes~\cite{caesar2020nuscenes}.
\end{itemize}

%% file: figures_tex/arch.tex
\begin{figure*}[t]
\begin{center}
 \includegraphics[width=1\linewidth]{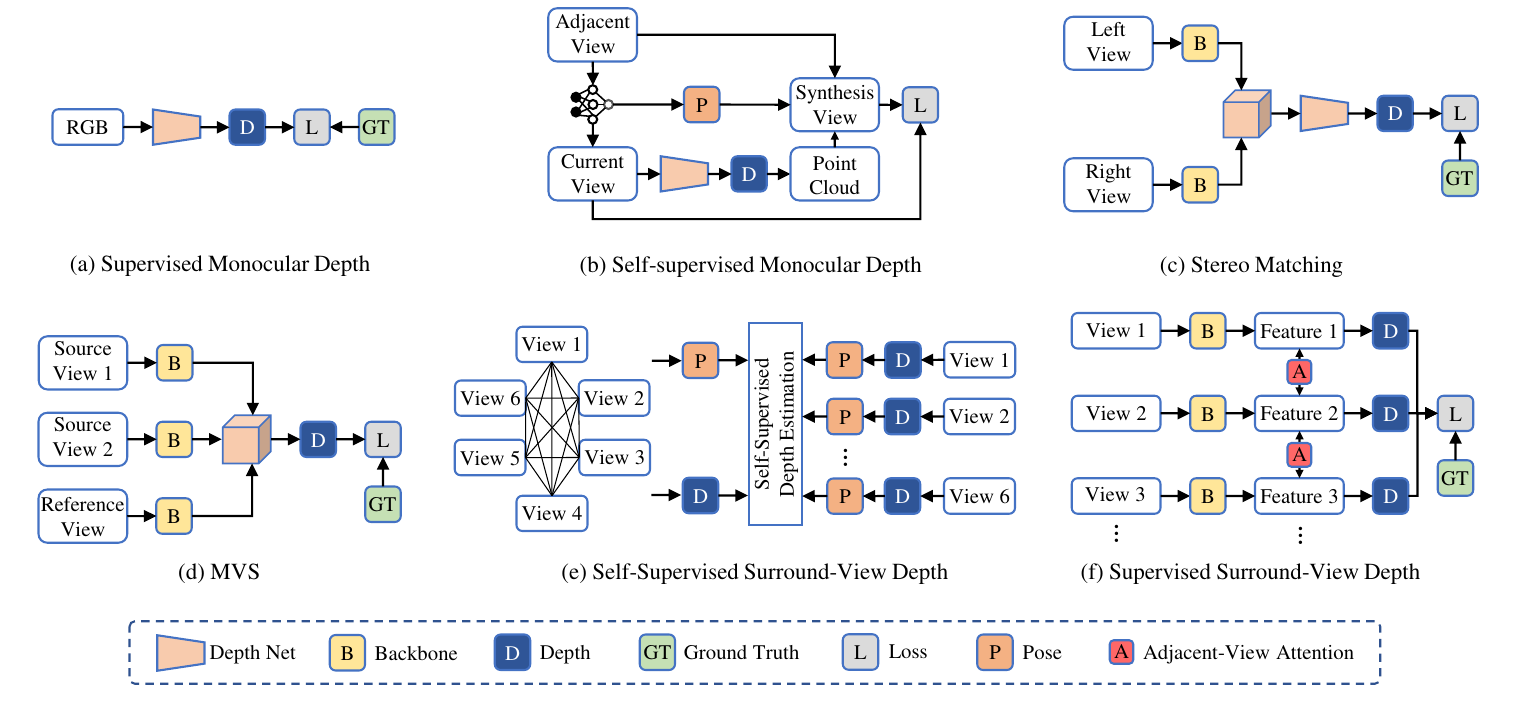}
\end{center}
\vspace{-0.5cm}
\caption{\textbf{Paradigm comparison} of Supervised Monocular Depth, Self-supervised Monocular Depth, Stereo Matching, Multi-View Stereo (MVS), Self-supervised Surround-View Depth and Supervised Surround-View Depth.}
\vspace{-0.55cm}
\label{fig:arch}
\end{figure*}

%% file: IROS/2relatedwork.tex
\section{Related Work}

\noindent \textbf{Monocular Depth Estimation.}
Monocular depth estimation has been studied in computer vision for a long time and plays a crucial role in 3D reconstruction, autonomous driving cars, and robot navigation. Supervised approaches take a single image and use datasets collected by laser scanners~\cite{kitti} or other range sensors~\cite{Silberman:ECCV12} as target depth labels in training, as shown in Fig.\ref{fig:arch}(a). Although learning-based depth estimation methods~\cite{eigen2015predicting,eigen2014depth,hu2019revisiting,li2022binsformer, PixelFormer, VA-Depth, Yuan_2022_CVPR, depthanything, depthanythingv2} achieve impressive results when compared with traditional machine learning baselines~\cite{saxena2008make3d,zhuo2015indoor,liu2014discrete}, these methods need additional sensors which are used to collect ``ground-truth" data. Besides, these methods are designed for the KITTI datasets~\cite{kitti} whose ``ground-truth" depth maps are dense. Moreover, the ``ground-truth" data may be sparse and not be very reliable due to high noise during data collection in practical applications. Thus, self-supervised  methods~\cite{zhou2017unsupervised,wang2017orientation,chang2018pyramid,yin2018geonet,monodepth2,bian2019unsupervised,zhou2019moving,ranjan2019competitive,zhao2020towards} explore the routes of jointly learning monocular depth and ego-motion from unlabeled sequential data, as shown in Fig.\ref{fig:arch}(b).
Inspired by the success of transformer~\cite{dosovitskiy2020image,liu2021swin,lee2021mpvit,ranftl2021vision,yang2021transformer,depthformer}, MonoVit~\cite{monovit} further improves depth estimation accuracy by combining plain convolutions with transformer blocks.  Recently, some works have aimed to build a powerful foundation model, such as MiDaS~\cite{midas}, DepthAnything~\cite{depthanything,depthanythingv2}, and Marigold~\cite{Marigold}.
However, these methods are designed for monocular depth estimation (typically less than $180^\circ$), consistent surround-view depth estimation would extend these approaches to the full $360^\circ$ space.

\noindent \textbf{Stereo Matching.}
Stereo matching predicts disparity between the left and right image, as shown in Fig.\ref{fig:arch}(c). In general, the stereo depth estimation task treats the 2D matching problem as a 1D disparity search. Traditional methods~\cite{sgm2005, sgm2007} utilize hand-crafted features to computer matching cost, which tends to fail on those textureless and repetitive regions in the images.  In recent years, with the support of large synthetic datasets \cite{mayer2016large, sintel, fallingthings}, the end-to-end network has achieved impressive performance on stereo matching. One line of networks~\cite{mayer2016large, pang2017cascade, liang2018learning, guo2019group, xu2020aanet, tankovich2021hitnet,wang2020fadnet} match features by dot-product correlation, which only uses 2D convolutions. Another line of networks~\cite{kendall2017end,chang2018pyramid,khamis2018stereonet,zhang2019ga, yang2019hierarchical, cheng2020hierarchical} builds 4D feature volume and learns to aggregate the matching cost by 3D convolutions. Besides, Li~\etal~\cite{Li_2021_ICCV} take advantage of the transformer architecture, which has advantages in feature matching and relaxes the limitation of a fixed disparity range, called STTR. However, a large overlap between the left and right images is needed for stereo matching. Our proposed method is intended for consistent multi-camera configurations with minimal image overlap.

\input{figures_tex/S3Depth}

\noindent \textbf{Surround-view Depth Estimation.}
Based on the way of supervision, surround-view depth estimation methods can be divided into self-supervised and supervised. To our knowledge, FSM~\cite{guizilini2021full} first extends self-supervised monocular depth estimation to a large-baseline multi-camera setting. 
Following FSM~\cite{guizilini2021full}, SurroundDepth~\cite{wei2022surround} recovers the real-world scales by pretraining the depth network with sparse depths obtained from SFM~\cite{schonberger2016structure} and learns to capture the surround-view interactions using self-attention layers, as shown in Fig.\ref{fig:arch}(e).
Recently, R3D3~\cite{schmied2023r3d3} leverages both spatial and temporal cues for dense 3D reconstruction from multiple cameras in a self-supervised manner.
For the task of supervised multi-view depth estimation, it is an effective method to build cost volume~\cite{mvsnet, wei2021nerfingmvs, xue2019mvscrf, huang2018deepmvs, tong2022normal}, namely, multi-view stereo (MVS). MVS aims to estimate the depth map by a series of calibrated images surrounding an object, and it is a one-to-many feature matching task, as shown in Fig.\ref{fig:arch}(d). Different from MVS, our model is designed to estimate all depth maps of surrounding multi-view images with small overlapping areas between cameras.

%% file: figures_tex/S3Depth.tex
\begin{figure*}[t]
\begin{center}
\includegraphics[width=1 \linewidth]{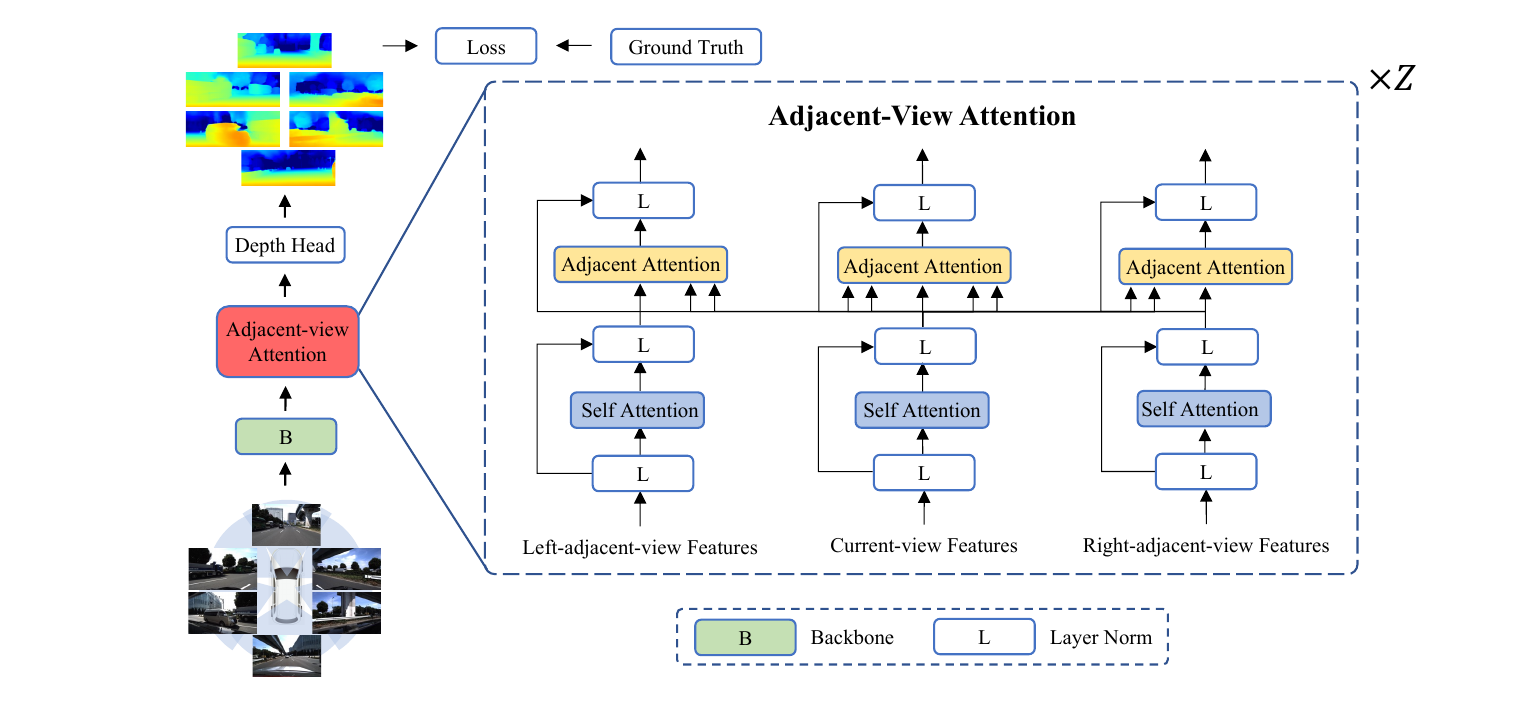}
\end{center}
\vspace{-0.5cm}
\caption{\textbf{The architecture of our proposed \net{}}. Our \net{} consists of three parts: Feature Extraction, Adjacent-view Attention, and Depth Head. For feature extraction, both transformer and convolutional layers are adopted to enhance the feature modeling and depth inferring. For Adjacent-view Attention, the alternating self-attention and adjacent-attention are adopted for $Z$ times to fully exchange the information between the current view and the adjacent view. Finally, Depth Head predicts the all-view depth map at the same time.}
 \vspace{-0.5cm}
\label{fig:S3Depth}
\end{figure*}

%% file: IROS/3method.tex
\section{Proposed Method} \label{method}

\subsection{Problem Formulation}

Given ${N}$ views input samples ${I} = \{{I}^{1}, {I}^{2}, \cdots  {I}^{N}\}$, the surround-view depth model $f(\cdot)$ is expected to predict the depth map  ${D}^{pred} = \{{D}^{pred}_{1}, {D}^{pred}_{2}, \cdots  {D}^{pred}_{N}\}$, where ${I}^{i} \in \mathbb{R}^{H \times W \times 3}$ denotes the surround-view RGB images, ${D}^{pred}_{i} \in \mathbb{R}^{H\times W}$  denotes the predicted depth map and ${H}$, ${W}$ stand for the height and width of input images, respectively. In \net{}, the only supervised signal comes from LiDAR. Different from monocular depth estimation, there are crucial overlaps across the surround-view images, which are very useful and can be imported for the model to understand the scene as a whole. Based on that, we introduce a simple baseline \net{} to predict $360^\circ$ view depth at the same time. By taking advantage of the adjacent-view information interactions, \net{} can achieve better performance of every view than view-dependent estimations.

\subsection{Overview}
Fig.\ref{fig:S3Depth} shows our proposed \net{} architecture. Our purpose is to build a powerful baseline for dense depth prediction of surround-view images, so we employ a multi-stage model~\cite{lee2021mpvit} to get enriched representations. Specifically, the baseline consists of three parts: Feature Extraction, Adjacent-view Attention, and Depth Head. (1) Feature extraction is a shared backbone that extracts 2D features from $N$ views of input images. (2) The adjacent-view Attention architecture is leveraged to exchange adjacent-view information of features. Specifically, the alternating attention mechanism is adopted: self-attention computes attention to features in the same view, while adjacent attention computes attention to features in the adjacent views to exchange the adjacent multi-scale representations at every scale. (3) Different from monocular depth estimation, the Depth Head predicts the surround-view depth map at the same time.

\subsection{Feature Extractor}

\textbf{Image Encoder.} Recently, the attention mechanism has already proven to be effective in various computer vision tasks, such as image classification \cite{liu2021swin}, object detection \cite{carion2020end} and depth estimation \cite{ranftl2021vision,yang2021transformer,depthformer,monovit,zhang2023completionformer}. Benefiting from the inductive bias, the transformer allows the model to focus on the important parts of the feature. Although the transformer can model global context perception, it may ignore local relationships. On the contrary, CNN can take advantage of local connectivity. Therefore, we adopt a module that combines CNN with transformer~\cite{lee2021mpvit} within \net{}, which can learn to interact between local and global features for enriched representations in five stages. 
Multiple surround-view images are concatenated and then fed into a feature extractor.
Specifically, in the first stage, a Conv-stem block is adopted to extract 2D features $F_{1} \in \mathbb{R}^{N \times C_F \times \frac{H}{2} \times \frac{W}{2}}$, where $H$ and $W$ indicate the height and width of images, $N$ represents views of input images and $C_F$ stands for the feature channel number. $F_{1}$ denotes features of multiple views. For the next four stages $i \in \{2,3,4,5 \}$, the joint convolution attention and transformer blocks are used to extract a features pyramid $\{F_{2}\in \mathbb{R}^{N \times C_F \times \frac{H}{2} \times \frac{W}{2}}, F_{3}\in \mathbb{R}^{N \times C_F \times \frac{H}{4} \times \frac{W}{4}}, F_{4}\in \mathbb{R}^{N \times C_F \times \frac{H}{8} \times \frac{W}{8}}, F_{5}\in \mathbb{R}^{N \times C_F \times \frac{H}{16} \times \frac{W}{16}}\}$, which can represent multi-scale features.

\textbf{Depth Decoder.} In the depth decoder, we fuse the multi-scale features of the same view and the features of cross-view images and gradually increase the spatial resolution. Then, the features are aggregated at $\frac{1}{2}$, $\frac{1}{4}$, $\frac{1}{8}$, $\frac{1}{16}$ resolutions respectively. Moreover, the output maps at $\frac{1}{2}$, $\frac{1}{4}$, $\frac{1}{8}$, $\frac{1}{16}$ resolutions are upsampled to the input resolution and then compute the loss function. Finally, a skip connection is utilized to connect input and output features to reduce the information loss of downsampled feature maps.

\subsection{Adjacent-view Attention}

The detailed structure of the adjacent-view attention architecture is provided in Fig.\ref{fig:S3Depth}. Different from SurroundDepth~\cite{wei2022surround}, which utilizes a cross-view transformer that combines six views through stitching and implements it using self-attention layers, we adopt the alternating self-attention and adjacent-attention. Self-attention extracts features within the current view and adjacent-attention is used to exchange representations across adjacent view features. Let $F_{ij} \in \mathbb{R}^{N \times C_F \times \frac{H}{n} \times \frac{W}{n}}, i=1,2,3,4,5,j=1,2,3,4,5,6$ be the feature maps obtained from the $j$-th view of the $i$-th stage, where $H$ and $W$ indicates the height and width of images, $N$ represents views of input images, $C_F$ stands for the feature channel number. For every stage, multi-head attention is used to increase the expressiveness of the features. The attention mechanism uses the dot product similarity to calculate the attention between the $query$ vectors and the $key$ vectors, and then weigh the $value$ vectors. For self-attention, the $\mathcal{Q},\mathcal{K},\mathcal{V}$ are computed from the same view using feature descriptors $F_{ij}$ as input, where $W_{\mathcal{Q}}, W_{\mathcal{K}}, W_{\mathcal{V}} \in \mathbb{R}^{C \times C}$, $b_{\mathcal{Q}}, b_{\mathcal{K}}, b_{\mathcal{V}} \in \mathbb{R}^{C}$ and C represents the embedding dimension respectively:
\begin{equation}
    Q_{i,j} = W_{\mathcal{Q}}F_{i,j} + b_{\mathcal{Q}}, K_{i,j} = W_{\mathcal{K}}F_{i,j} + b_{\mathcal{K}}, V_{i,j} = W_{\mathcal{V}}F_{i,j} + b_{\mathcal{V}},
\end{equation}
\begin{equation}
    \begin{split}
        F^{sf}_{i,j} &= \text{softmax}(\frac{Q_{i,j}^T K_{i,j}}{\sqrt{C}}){V_{i,j}}.
    \end{split}
\end{equation}

For adjacent-attention, $\mathcal{K},\mathcal{V}$ are computed from the current-view feature, while $\mathcal{Q}$ is computed from the adjacent-view feature, where $x$ refers to adjacent views ($x \in {j-1, j+1} $). Specifically, when $j$ equals 1,  $x$ equals 6 and 2; Similarly, when $j$ equals 6, $x$ equals 5 and 1: 

\begin{equation}
    Q_{i,x} = W_{\mathcal{Q}}F^{sf}_{i,x} + b_{\mathcal{Q}},
\end{equation}
\begin{equation}
    K_{i,j} = W_{\mathcal{K}}F^{sf}_{i,j} + b_{\mathcal{K}}, 
\end{equation}
\begin{equation}
    V_{i,j} = W_{\mathcal{V}}F^{sf}_{i,j} + b_{\mathcal{V}},
\end{equation}

\begin{equation}
    \begin{split}
    F^{adj}_{i,j} &= \text{softmax}(\frac{Q_{i,x}^T K_{i,j}}{\sqrt{C}}){V_{i,j}}.
    \end{split}
\end{equation}

To make information fully interactive between the current view and adjacent views, the alternating self-attention and adjacent-attention are used for $Z$ times.

\subsection{Depth Head.}

To achieve a comprehensive understanding and powerful representation features at the same location at different scales are used together to predict the final depth. By taking advantage of the adjacent-view attention, features can interact with adjacent views to compensate for erroneous estimations. The final output can be represented as follows:
\begin{equation}
    \begin{split}
        D^{pred}_{i} &= \sigma(F^{out}_{i}){D_{max}},
    \end{split}
\end{equation}
where $\sigma$ stands for the $sigmoid$ activation function and $D_{max}$ denotes the max distance. $F^{out}_{i}$ represents the upsampled features of the adjacent attention from each stage, bringing them back to the original image resolution.

\subsection{Loss Function}\label{loss_function}

We directly use point cloud data as supervision for the depth estimation. Specifically, the point cloud is projected onto the image plane. $L_1$ loss is adopted to calculate the distance with the ground-truth depth on the valid pixels as follows:
\begin{equation}
    L_{depth} = \frac{1}{N_{valid}} \sum_i\mathbb({D}^{gt} > 0)\left|{D}^{pred}_{i} - {D}^{gt}_{i}\right|,
\end{equation}

Inspired by recent self-supervised depth estimation \cite{monodepth2} methods, an L1 penalty on the depth map is used to encourage local smoothness. 
\begin{equation}
L_{smooth} = \sum_x  \left|\partial_x d'_x\right|e^{-\left\|\partial_x I_x \right\|} + \left| \partial_y d'_x \right| e^{-\left\| \partial_y I_x\right\|}.
\end{equation}

Finally, we combine $L_{depth}$ and $L_{smooth}$ to supervise the network: 
\begin{equation}
L = L_{depth} + \lambda L_{smooth}.
\end{equation}

%% file: IROS/4experients.tex
\section{Experiments}

\subsection{Datasets and Evaluation Metrics}

\textbf{NuScenes \cite{caesar2020nuscenes}}. 
The nuScenes benchmark is an up-to-date dataset for 3D object detection, as well as semantic segmentation. 
It collects 1000 scenes which are divided into 700/150/150 scenes for training/validation/testing. The ground truth depths are generated by processing the LiDAR point cloud. During the training and evaluation procedure, we consider the max distance $D_{max}$ as 80m.

\textbf{DDAD \cite{3Dpacknet}}. 
The DDAD dataset is a large-scale dataset for long-range (up to 250m) and dense depth estimation in challenging conditions. 
There are 12,650 training samples (75,900 images) and 3,950 validation samples (15,800 images) in this dataset. During the training and evaluation procedure, we consider the max distance $D_{max}$ as 200m.

\textbf{Evaluation Metrics}.
We assess the results of our method following the evaluation protocol of the depth estimation task~\cite{monodepth2, 3Dpacknet, guizilini2021full, wei2022surround}. 
While evaluating self-supervised depth estimation methods, the scores are averaged per-frame with median-scaling~\cite{guizilini2021full, wei2022surround}.

\subsection{Implementation Details}

\textbf{BTS}~\cite{BTS} We use the official implementation of BTS~\cite{BTS}. Densenet161~\cite{huang2017densely} is used as the backbone. The number of epochs is set to 10 for nuScenes and DDAD. 

\textbf{PixelFormer}~\cite{PixelFormer} 
We use the official implementation of PixelFormer~\cite{PixelFormer}. Swin-Tiny~\cite{liu2021swin} is used as the backbone. The number of epochs is set to 10 for nuScenes and 20 for DDAD.

\textbf{SurroundDepth}~\cite{wei2022surround} We use the official implementation of SurroundDepth~\cite{wei2022surround}, and only change the supervision from self-supervised to LiDAR supervised. 

\textbf{\net{}}
We conduct experiments on both DDAD~\cite{3Dpacknet} and nuScenes~\cite{caesar2020nuscenes} datasets for depth estimation. We employ MPViT-small with ImageNet \cite{deng2009imagenet} pre-trained weight as the encoder. At each scale, we use $Z=8$ adjacent-view attention layers. Eight NVIDIA 3090 24GB GPUs are used to train our model. The value of $\lambda$ is set to 0.01 in this paper. Adam \cite{kingma2014adam} is employed as our optimizer with $\beta_1 = 0.9$ and $\beta_2 = 0.999$ on both datasets. For the nuScenes~\cite{caesar2020nuscenes} datasets, we train the \net{} with batch size 48, learning rate 1e-4 for 10 epochs. The resolution of input images is downsampled to 640$\times$352. When evaluating the models, the resolution of the predicted depth map is upsampled to 900$\times$1600. For the DDAD dataset which is larger than nuScenes, we train the \net{} with batch size 48, learning rate 1e-4 for 20 epochs. The resolution of input images is downsampled to $640 \times 384$ on DDAD. During the evaluation, the resolution of the predicted depth map is upsampled to $1216 \times 1936$.

\input{table_tex/nusc_s}

\subsection{Comparison to State-of-the-Art}

\textbf{Results on nuScenes}.
Because few works study the task of supervised surround-view multi-camera depth estimation, we build a benchmark on the nuScenes~\cite{caesar2020nuscenes} dataset, as shown in~\autoref{tab:nusc}.
Specifically, we change the supervision of SurroundDepth~\cite{wei2022surround} from self-supervised to LiDAR supervised and get the result from 0.245 to 0.158 in Abs Rel. \net{} respectively reaches $0.067$ Abs Rel, which exceeds the recently proposed methods: SurroundDepth~\cite{wei2022surround} (Abs Rel $0.158$) and PixelFormer~\cite{PixelFormer} (Abs Rel $0.113$). This demonstrates that \net{} can further improve the performance of surround-view depth estimation even while SurroundDepth~\cite{wei2022surround} and PixelFormer~\cite{PixelFormer} have achieved impressive performance.

\input{table_tex/ddad_s}

\input{table_tex/cross-data}

\textbf{Results on DDAD}.
We also train and test our model in DDAD~\cite{3Dpacknet} datasets, which are known for featuring denser depth estimation in challenging conditions when compared to the nuScenes~\cite{caesar2020nuscenes} datasets. As indicated in \autoref{tab:depth}, our proposed \net{} also achieves state-of-the-art performance across all evaluation metrics.
From these results, it is evident that our proposed approach, \net{}, is capable of predicting more accurate object structures than other competing methods. Our findings suggest that \net{} can improve the accuracy of depth estimation in challenging conditions and has the potential to be applied in various real-world scenarios.

\textbf{Cross-dataset transfer}.
Compared to previous methods, \net{} achieves the lowest Abs Rel (0.292), Sq Rel (4.201), and RMSE (7.481) while attaining the highest $\delta 1$ accuracy (0.665). Notably, it outperforms R3D3~\cite{schmied2023r3d3}, the previous best-performing method, particularly in Sq Rel and RMSE, demonstrating improved generalization across datasets. These results highlight the effectiveness of \net{} in adapting to domain shifts, ensuring more reliable depth estimation in diverse driving scenarios.

\input{table_tex/ablation_cvt}

\input{figures_tex/qualitative} 

\subsection{Ablation Study}

\textbf{Effect of input}. 
The first two rows of the \autoref{tab:ablation_cvt} show the effect of different inputs. It yields better results to use surround-view images as input, compared to using random images. This is because surround-view images have overlapping regions which can stabilize gradients, resulting in improved performance. Specifically, the depth estimation results are consistent in the overlap regions between views, which provide additional information that can improve the accuracy of the model's predictions.  
Moreover, this overlapping information can also help the model to better handle changes in perspective, rotation, and scale.

\textbf{Effect of adjacent-view attention}. 

As shown in \autoref{tab:ablation_cvt}, the proposed adjacent-attention mechanism facilitates the interaction between adjacent-view information, thereby enabling adjacent-view features to interact with each other. This demonstrates the efficacy of the adjacent-view attention module in enhancing the performance of our model across different backbones. This module effectively enables the network to learn to interact between current and adjacent view features. 
Moreover, the adjacent-view attention can not only improve the performance but also accelerate the convergence of the model, as shown in Fig.\ref{fig:train_loss_iters}. 
\input{figures_tex/train_loss_iters}

\input{table_tex/ablation_mpvit}
\input{table_tex/ablation_Z}
\input{table_tex/ablation_loss}

\textbf{Effect of backbone}. 
To delve into the effectiveness of different backbones, we report an ablation study on the nuScenes~\cite{caesar2020nuscenes} val set, as shown in \autoref{tab:ablation_mpvit}. 
Also, we report the flops and parameters of different backbones. Benefiting from the module combining CNN and Transformer layers, the MPViT~\cite{lee2021mpvit} backbone achieves better performance than pure Transformer backbones (PVT-small~\cite{wang2021pyramid}), the pure CNN backbone (ResNet34~\cite{resnet}), and hybrid models that combine convolutional networks with transformers (CVT~\cite{wu2021cvt}). In addition, we further explore the effects of different modules in MPViT~\cite{lee2021mpvit}. As shown in the bottom of \autoref{tab:ablation_mpvit}, both the CNN path and the Transformer path play an important role in MPViT~\cite{lee2021mpvit} for depth estimation. 

\textbf{Effect of adjacent-view attention layers}. 
We also conduct an ablation study to analyze the impact of the number of adjacent-view attention layers, as shown in \autoref{tab:ablation_Z}. The model achieves its best results when $Z = 8$, highlighting the significance of carefully selecting the number of attention layers

\textbf{Effect of loss}. 
As shown in \autoref{tab:ablation_loss}, we examine the impact of loss on the performance of our \net{}. Remarkably, our results indicate that our \net{} is highly robust to different types of loss functions. 

%% file: table_tex/nusc_s.tex
\begin{table}[t]
	\centering
         \caption{\textbf{Results on the nuScenes\cite{caesar2020nuscenes} \emph{val} set.} $^*$ indicates the implementation of SurroundDepth~\cite{wei2022surround}. The symbol $^\dagger$ indicates our implementations. \emph{S} denotes self-supervised, \emph{D} denotes depth supervision with LiDAR and \emph{MS} indicates that the scores are averaged per-frame with median-scaling~\cite{guizilini2021full, wei2022surround} at test time.}.
	\resizebox{0.5\textwidth}{!}{
		\begin{tabular}{l|c|c||cccc}
		    \hline
            Method&Train&Test&Abs Rel$\downarrow$& Sq Rel$\downarrow$&  RMSE$\downarrow$ & $\delta 1 \uparrow$ \\
            \hline 
			Monodepth2$^*$ \cite{monodepth2} &  S & MS & 0.287  &   3.349  &   7.184   &   0.641  
			\\
			
			PackNet-SfM$^*$ \cite{3Dpacknet} &  S &MS &   0.309  &   2.891  &   7.994  &      0.547 
			
			\\
			FSM \cite{guizilini2021full} &  S & MS  &   0.299  &   -  &   -  &   - 
			
			\\
			FSM$^*$ \cite{guizilini2021full} &  S & MS  &   0.334  &   2.845  &   7.786    &   0.508  
			
			\\
			SurroundDepth$^*$ \cite{wei2022surround} &  S & MS  &   0.245  &  3.067  & 6.835  &  0.719  
			\\
            MCDP~\cite{xu2022multi} &  S  &MS&0.237& 3.030 &6.822 &0.719\\
S3D3~\cite{schmied2023r3d3} &S&MS&0.235 &3.332& 6.021 &0.749\\
			\hline
			BTS$^\dagger$ \cite{BTS} &  D  & - &  0.128&  0.788&  3.750&    0.862

                \\
			PixelFormer$^\dagger$ \cite{PixelFormer}&  D  & - &  0.099&	0.553&	3.171	&	0.904	
                \\
               
                \hline
                SurroundDepth$^\dagger$ \cite{wei2022surround} &  D  & - & 0.158  & 1.189  & 5.028    & 0.813  
                \\

			\textbf{Ours}  & D & - & \bf{0.067}  & \bf{0.673}  & \bf{2.457} & \bf{0.951}
			\\
			\hline
	\end{tabular}
 }
	\vspace{-0.5cm}
	\label{tab:nusc}
\end{table}


%% file: table_tex/ddad_s.tex
\begin{table}[t]
	\centering
         \caption{\textbf{Results on DDAD \cite{3Dpacknet}.} \emph{-M} indicates occlusion masking. 
        }
	\resizebox{0.5\textwidth}{!}{
    \begin{tabular}{l|c|c||cccc} 
        \toprule
        Method&Train&Test&Abs Rel$\downarrow$& Sq Rel$\downarrow$&  RMSE$\downarrow$ & $\delta 1 \uparrow$ \\
        \midrule 
    Monodepth2$^*$ \cite{monodepth2}   & S & MS&  0.362  &  14.404  &  14.178  &   0.683      
 
    \\
    PackNet-SfM$^*$ \cite{3Dpacknet}  & S & MS&  0.301  &   5.339  &  14.115  &      0.624  
    
    \\			
    Monodepth2 -M$^*$    & S & MS & 0.217  &   3.641  &  12.962  &     0.699  
    \\
    PackNet-SfM -M$^*$   & S & MS& 0.234  &   3.802  &  13.253  &   0.672 
    \\
    FSM \cite{guizilini2021full}  & S & MS & 0.202  &   -  &  -  &  -  
    
    \\
    FSM$^*$ \cite{guizilini2021full}  & S & MS & 0.229  &   4.589  &  13.520   &   0.677  
    
    \\
    SurroundDepth$^*$ \cite{wei2022surround}  & S & MS & 0.200 &   3.392  &  12.270  &  0.740 
    \\
    MCDP~\cite{xu2022multi}& S&MS& 0.237 &3.030& 6.822 &0.719
    \\
    S3D3~\cite{schmied2023r3d3} &S&MS & 0.169& 3.041& 11.372 &0.809
    \\
    \hline
    BTS$^\dagger$ \cite{BTS}  & D  & - &  0.204&  3.118&  12.330&   0.694
        \\
    
    PixelFormer$^\dagger$ \cite{PixelFormer}  & D  & - &  0.177&	2.525&	11.048	&0.747
        \\
    
    \midrule
    SurroundDepth$^\dagger$ \cite{wei2022surround}  & D  & -& 0.175  & 2.694  & 11.607  & 0.757  
    \\
    \textbf{Ours} &  D & - &   \bf{0.160}  &  \bf{2.527}  & \bf{10.803}  &   \bf{0.799}  
    \\
    \bottomrule
\end{tabular}
}
\vspace{-0.5mm}
\label{tab:depth}
\end{table}

%% file: table_tex/cross-data.tex
\begin{table}[t]
	\centering
         \caption{\textbf{Cross-dataset transfer}. Models trained on DDAD are evaluated for scale-aware depth prediction on the NuScenes dataset. 
        }
	\resizebox{0.5\textwidth}{!}{
\begin{tabular}{l|cccc} 
\toprule
Method &Abs Rel$\downarrow$& Sq Rel$\downarrow$&  RMSE$\downarrow$ & $\delta 1 \uparrow$   \\
\midrule
FSM~\cite{guizilini2021full}  & 0.349 & 5.064 & 8.785 & 0.499 \\
SurroundDepth~\cite{wei2022surround}  & 0.364 & 5.476 & 8.447 & 0.525 \\
R3D3~\cite{schmied2023r3d3} & 0.292 & 4.800 & 7.677 & 0.660 \\
\textbf{Ours}&  \textbf{0.292}  &   \textbf{4.201}  &   \textbf{7.481}&   \textbf{0.665} \\
\bottomrule
\end{tabular}
}
\end{table}

%% file: table_tex/ablation_cvt.tex
\begin{table}[t]
\centering
\caption{\textbf{Ablation study on nuScenes~\cite{caesar2020nuscenes} \emph{val} set. -- Input and Adj.} \emph{Adj.} stands for Adjacent-view Attention. \emph{Random} indicates that the input images come from diverse time frames. \emph{Surr.} denotes surround-view images captured at the same time.}
\scalebox{0.58}{
\begin{tabular}{l|c|c||cccc|ccc}
\toprule

\multirow{2}{*}{Model} &\multirow{2}{*}{Input} &\multirow{2}{*}{+Adj.} &\multicolumn{4}{c|}{lower is better} & \multicolumn{3}{c}{higher is better}\\
&& &Abs Rel$\downarrow$& Sq Rel$\downarrow$&  RMSE$\downarrow$ & RMSE log $\downarrow$& $\delta_{1} \uparrow$ & $\delta_{2} \uparrow$ & $\delta_{3} \uparrow$\\
\midrule
ResNet34~\cite{resnet}&Random&  & 0.091  &0.861  &3.105  &0.178 &0.927  &0.957  &0.972 \\
 ResNet34~\cite{resnet}&Surr.&& 0.080 & 0.842 & 2.882 & 0.166 & 0.937 & 0.961 &0.975 \\
ResNet34~\cite{resnet}&Surr.&\checkmark&  0.074& 0.755&	2.708& 0.157&	0.945& 0.964& 0.976 \\
\midrule
MPViT~\cite{lee2021mpvit} &Surr.&& 0.075&	0.718&	2.703&	0.158&	0.943&	0.964&	0.976 \\
MPViT~\cite{lee2021mpvit}&Surr.&\checkmark & \bf{0.067}  & \bf{0.673}  & \bf{2.457} & \bf{0.144} & \bf{0.951} & \bf{0.970} & \bf{0.981}  \\
\bottomrule
\end{tabular}}
\label{tab:ablation_cvt}
\vspace{-0.35cm}
\end{table}

%% file: figures_tex/qualitative.tex
\begin{figure*}[t]
\begin{center}
 \includegraphics[width=0.88\linewidth]{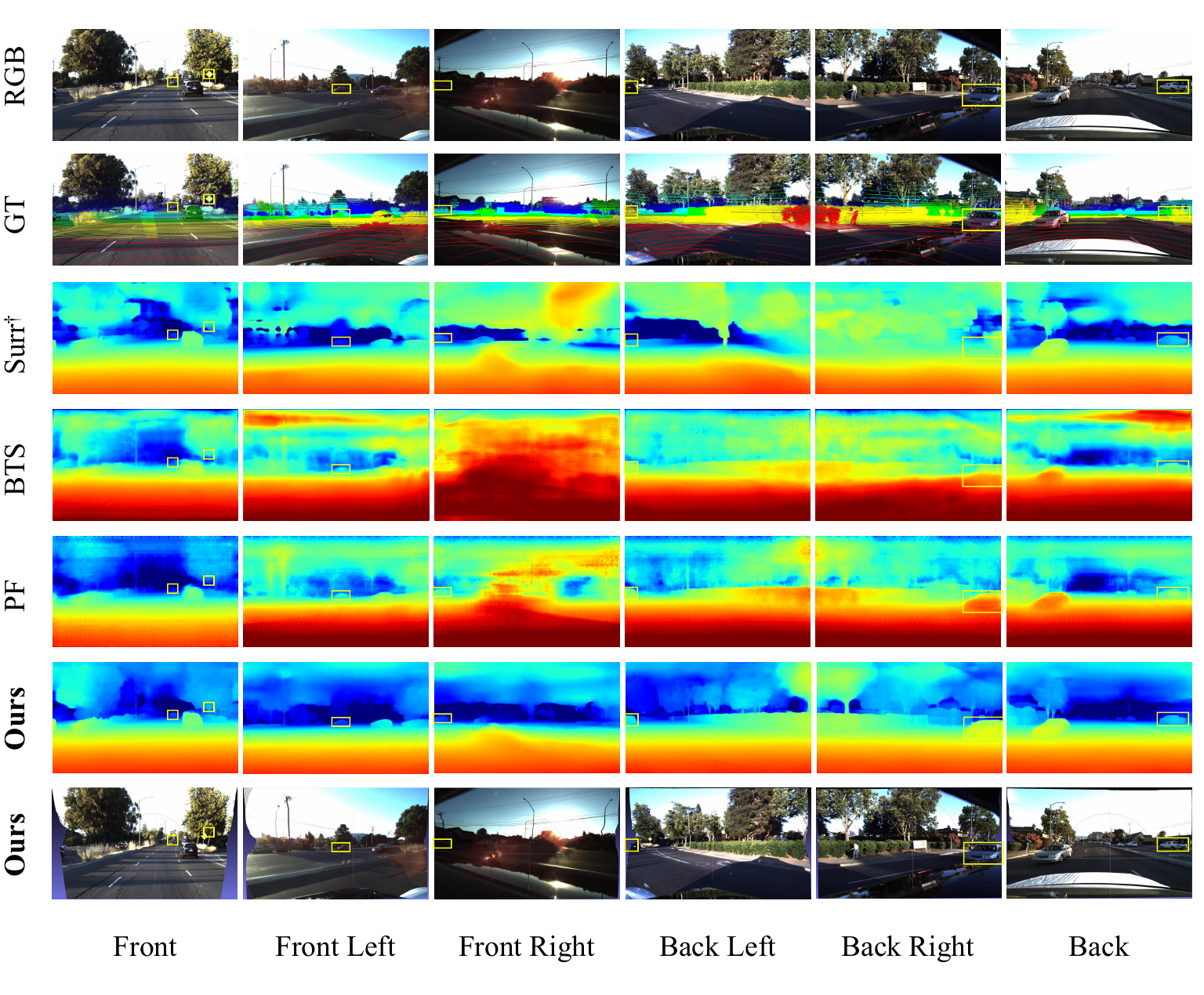}
\end{center}
\vspace{-0.6cm}
\caption{\textbf{Qualitative results on DDAD \cite{3Dpacknet} dataset.} The top two rows are input images and the GT depth map. Then, the predicted depth maps are provided. The last row is RGB point cloud visualization. Surr$^\dagger$ refers to SurroundDepth~\cite{wei2022surround} supervised with LiDAR. PF denotes PixelFormer~\cite{PixelFormer}.}
\label{fig:qualitative}
\vspace{-0.35cm}
\end{figure*}

%% file: figures_tex/train_loss_iters.tex
\begin{figure}[t]
\begin{center}
 \includegraphics[width=0.6 \linewidth]{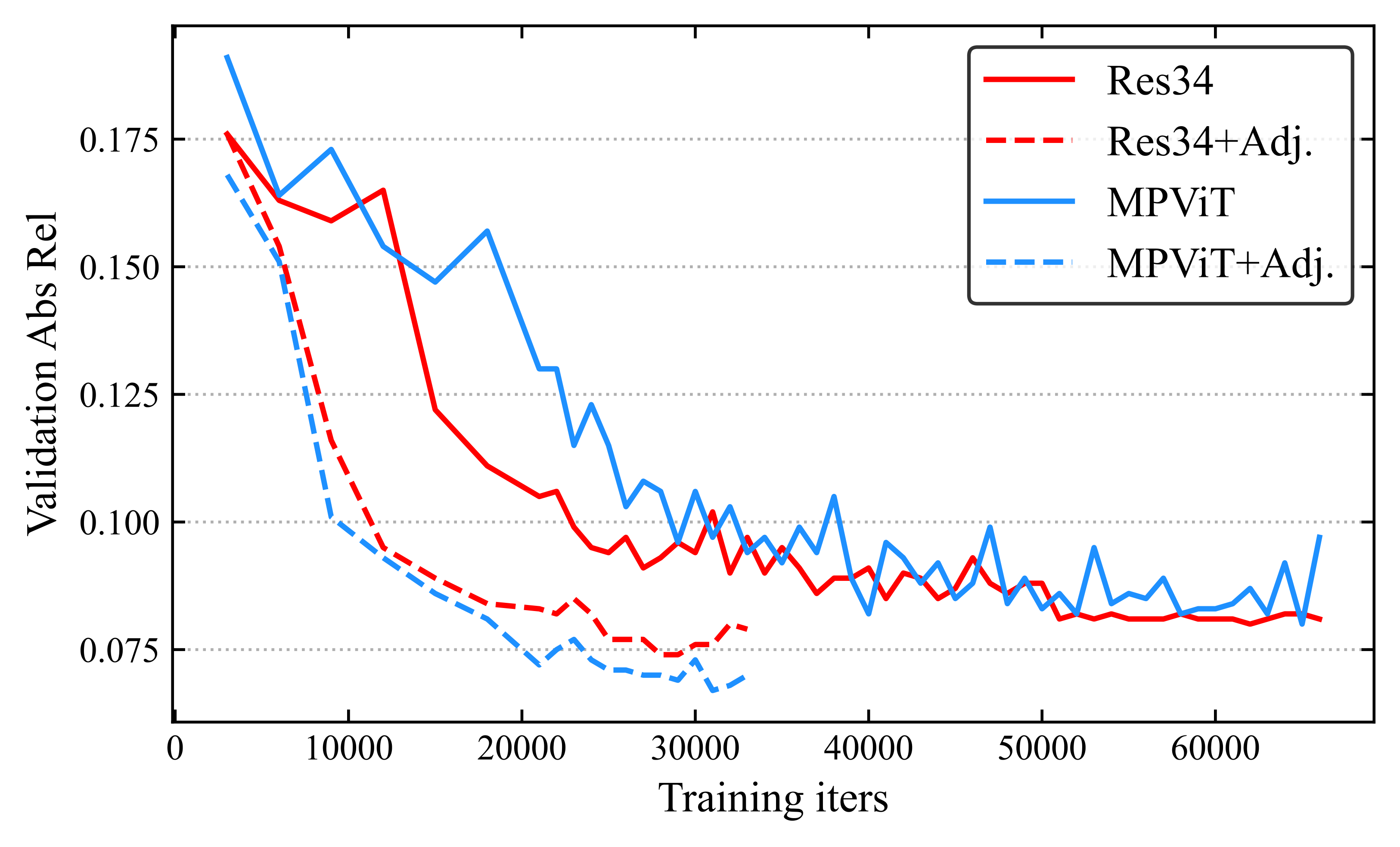}
\end{center}
\vspace{-0.3cm}
\caption{\textbf{Plot of Abs Rel in the nuScenes~\cite{caesar2020nuscenes} \emph{val} set.} Adj. stands for Adjacent-view Attention.}
\vspace{-0.4cm}
\label{fig:train_loss_iters}
\end{figure}

%% file: table_tex/ablation_mpvit.tex
\begin{table}[t]
\centering
\caption{\textbf{Ablation study on nuScenes~\cite{caesar2020nuscenes} \emph{val} set -- Backbone.} 
The input size is 640$\times$192, and the backbones are pre-trained on ImageNet~\cite{deng2009imagenet}. 
All backbones are added Adjacent-view Attention module.}
\scalebox{0.58}{
\begin{tabular}{l|cc||cccc|ccc}
\toprule
\multirow{2}{*}{Backbone}& \multirow{2}{*}{Flops} & \multirow{2}{*}{Params}&\multicolumn{4}{c|}{lower is better} & \multicolumn{3}{c}{higher is better}\\
 & & &Abs Rel$\downarrow$& Sq Rel$\downarrow$&  RMSE$\downarrow$ & RMSE log $\downarrow$& $\delta_{1} \uparrow$ & $\delta_{2} \uparrow$ & $\delta_{3} \uparrow$\\
\midrule 
ResNet34~\cite{resnet} &99.19G & 27M &  0.074& 0.755&	2.708& 0.157&	0.945& 0.964& 0.976  \\
PVT-small\cite{wang2021pyramid} &139.14G&7.9M&  0.127&	1.135&	3.842&	0.214&	0.879&	0.927&	0.956\\
CVT$^\dagger$ \cite{wu2021cvt}&170.73G&20M &0.083  &   0.781  &   2.896  &   0.164  &   0.936  &   0.961  &   0.975 \\

MPViT-tiny~\cite{lee2021mpvit} &49.53G& \textbf{10M} &  0.075&	0.787&	2.675&	0.159&	0.944&	0.964& 0.976 \\
MPViT-xsmall~\cite{lee2021mpvit} &85.04G&13M &   0.069&	0.731&	2.534&	0.150&	0.950&	0.968&	0.979 \\
MPViT-small~\cite{lee2021mpvit} &134.30G&27M & 
\bf{0.067}  & \bf{0.673}  & \bf{2.457} & \bf{0.144} & \bf{0.951} & \bf{0.970} & \bf{0.981}  \\

\midrule
 MPViT-small~\cite{lee2021mpvit} &134.30G&27M &    \bf{0.067}  & \bf{0.673}  & \bf{2.457} & \bf{0.144} & \bf{0.951} & \bf{0.970} & \bf{0.981}  \\
2 Trans. Path &85.04G&10.3M& 0.079&	0.814&	2.726&	0.163&	0.940&	0.962&	0.975 \\
1 Trans. Path &82.12G&11.2M&  0.076&	0.741&	2.757&	0.163&	0.940&	0.961&	0.974\\
w/o Trans. Path   &57.16G&11.5M& 0.082&	0.840&	2.881&	0.170&	0.936&	0.957&	0.971\\
\bottomrule
\end{tabular}
}
\label{tab:ablation_mpvit}
\vspace{-0.35cm}
\end{table}

%% file: table_tex/ablation_Z.tex
\begin{table}[t]
\centering
\caption{\textbf{Ablation study on nuScenes~\cite{caesar2020nuscenes} \emph{val} set -- $Z$.} $Z$ stands for the number of adjacent-view attention layers.}
\begin{tabular}{l|c c c c}
\toprule
\multirow{2}{*}{$Z$} &\multicolumn{4}{c}{lower is better} \\
&Abs Rel$\downarrow$& Sq Rel$\downarrow$&  RMSE$\downarrow$ & RMSE log $\downarrow$\\
\midrule 
4  & 0.071  &   0.757  &   2.507  &   0.148 	 \\
6  &  0.074  &   0.748  &   2.489  &   0.147\\
8 &\bf{0.067}  & \bf{0.673}  & \bf{2.457} & \bf{0.144} \\
10 &  0.071  &   0.729  &   2.532  &   0.149 \\
\bottomrule
\end{tabular}
\label{tab:ablation_Z}
\vspace{-0.5cm}
\end{table}


%% file: table_tex/ablation_loss.tex
\begin{table}[t]
\centering
\caption{Ablation study of loss on nuScenes~\cite{caesar2020nuscenes} \emph{val} set.}
\scalebox{0.88}{
\setlength
\tabcolsep{8pt}{
\footnotesize
\begin{tabular}{l |c c c c}
\hline

\multirow{2}{*}{Method} &\multicolumn{4}{c}{lower is better} \\
&Abs Rel$\downarrow$& Sq Rel$\downarrow$&  RMSE$\downarrow$ & RMSE log $\downarrow$\\
\hline 
L1 loss  &0.067&	0.632&	2.480&	0.145 \\
L1+Smooth loss &0.067&	0.673&	2.457&	0.144 \\
Silog loss  &0.069&	0.502&	2.434&	0.135 \\
Silog+Smooth loss &0.068&	0.485&	2.430&	0.134  \\
\hline
\end{tabular}}
}

\label{tab:ablation_loss}
\vspace{-0.2cm}
\end{table}

%% file: IROS/5conclusion.tex
\section{Conclusion}

In this paper, we build a baseline for supervised surround-view depth estimation. Different from monocular depth estimation, the core insight of \net{} is to exchange adjacent-view information and jointly predict all depths of surrounding views. 
We conduct extensive and detailed experiments comparing different backbones. 
Further, adjacent-view attention is employed at multiple and different scales to incorporate adjacent-view features. Through extensive experiments, we show that \net{} can achieve state-of-the-art performance on multi-camera depth estimation datasets.
We hope the proposed method can serve as a simple and strong baseline for multi-camera depth estimation.

\section*{ACKNOWLEDGMENT}
This work was supported by the National Natural Science Foundation of China under Grant 62373356 and the Joint Funds of the National Natural Science Foundation of China under U24B20162.